\documentclass[11pt,a4paper]{article}
\usepackage[hyperref]{acl2021}
\usepackage{times}
\usepackage{latexsym}

\usepackage{graphicx} 
\usepackage{amsmath}
\usepackage{multirow}
\usepackage{multicol}
\usepackage{mathrsfs}
\usepackage{booktabs}

\usepackage{microtype}

\aclfinalcopy 

\newenvironment{myitemize2}[1][]{
\begin{list}{$\bullet$}
    {
     \setlength{\leftmargin}{5mm}     
     \setlength{\parsep}{0.5mm}         
     \setlength{\topsep}{0mm}         
     \setlength{\itemsep}{0mm}        
     \setlength{\labelsep}{0.5em}     
     \setlength{\itemindent}{0mm}    
     \setlength{\listparindent}{6mm} 
    }}
{\end{list}}

\title{E2E-VLP: End-to-End Vision-Language Pre-training Enhanced by Visual Learning
}

\author{Haiyang Xu$\thanks{corresponding author}$,\ \  Ming Yan,\ \  Chenliang Li,\ \  Bin Bi,\ \ \\ \textbf{Songfang Huang,}\ \  \textbf{Wenming Xiao,}\ \  \textbf{Fei Huang}\ \ \\
Alibaba Group\ \ \\
\{shuofeng.xhy, ym119608, lcl193798, b.bi\}@alibaba-inc.com\ \ \\
\{songfang.hsf, wenming.xiaowm, f.huang\}@alibaba-inc.com}

\date{}

\begin{document}
\maketitle
\begin{abstract}
Vision-language pre-training (VLP) on large-scale image-text pairs has achieved huge success for the cross-modal downstream tasks. The most existing pre-training methods mainly adopt a two-step training procedure, which firstly employs a pre-trained object detector to extract region-based visual features, then concatenates the image representation and text embedding as the input of Transformer to train. However, these methods face problems of using task-specific visual representation of the specific object detector for generic cross-modal understanding, and the computation inefficiency of two-stage pipeline.


In this paper, we propose the first end-to-end vision-language pre-trained model for both V+L understanding and generation, namely E2E-VLP, where we build a unified Transformer framework to jointly learn visual representation, and semantic alignments between image and text. We incorporate the tasks of object detection and image captioning into pre-training with a unified Transformer encoder-decoder architecture for enhancing visual learning. 
An extensive set of experiments have been conducted on well-established vision-language downstream tasks to demonstrate the effectiveness of this novel VLP paradigm.
\end{abstract}

\section{Introduction}
Self-supervised pre-training has achieved great success in a wide range of natural language understanding ~\citep{devlin2018bert, liu2019roberta, wang2019structbert, lan2019albert} and generation tasks~\citep{song2019mass, lewis2019bart, bi2020palm}. Recent studies ~\citep{li2019visualbert, lu2019vilbert, chenuniter, tan2019lxmert, li2020oscar, yu2020ernie} have also witnessed the progress of self-supervised pre-training on vision-and-language tasks, which learns general cross-modal representations from massive image-text pairs, and fine-tunes vision-language pre-training (VLP) models on task-specific data achieving state-of-the-art results on various downstream V+L tasks.

Most existing mainstream VLP models adopt a two-step training method, which firstly extracts semantic visual features using a pre-trained object detection model, and then combines the derived object-centric representation of the image and text embedding as the input of Transformer~\citep{vaswani2017attention} for cross-modal pre-training. Despite the superior performance brought by the large-scale image-text pairs, the two-stage solution suffers from the following weaknesses: 1) the object detection model in the first step is trained on specific visual dataset such as Visual Genome dataset ~\citep{krishna2017visual}, and the visual representation is not optimized towards a more generic cross-modal understanding in the second step. It may suffer from an error propagation problem when the object detection model fails to recognize certain important information. 2) extracting region features with an object detection model is so time-consuming that most state-of-the-art models are directly trained and evaluated on cached visual features. This practice not only imposes unnecessary constraints on model designs, but also confronts the run-time inference inefficiency in the prediction phase.

Recently, several studies such as~\citep{jiang2020defense} have begun to revisit the grid features for cross-modal understanding and found the grid features can also work surprisingly well, while making the model design and training process much simpler. One pioneering work Pixel-BERT ~\citep{huang2020pixel} explores to pre-train with grid features in an end-to-end fashion directly from pixels. It removes all the fine-grained visual pre-training tasks, which proves to be important for V+L pre-training. 
~\citep{zhang2021vinvl} also demonstrates that visual features provided by the object detection model matter significantly in VLP models.

To address the limitations, we propose a new end-to-end paradigm for pixel-level vision-language pre-training, namely E2E-VLP, by enhancing with fine-grained visual learning. During pre-training, E2E-VLP jointly learns the visual region features and the cross-modal representation in a unified Transformer encoder-decoder architecture directly from image pixels. In addition to the typical pre-training tasks of Masked Language Modeling and Image-Text Matching, we enhance the vision-language pre-training with fine-grained visual semantic learning. Specifically, two end-to-end  pre-training tasks are further incorporated: 1) \textit{Object Detection}: inspired from DETR~\citep{carion2020end}, we view the object detection as a direct set prediction problem. The cross-modal Transformer encoder and image encoder are joint learnt to fuse the cross-modal data from pixels, while the decoder is used to capture fine-grained visual information via bipartite matching between predicted and ground-truth objects; 2) \textit{Image-Text Generation}: to better understand the semantics within the image, we also use the paired text to guide the learning of image features. We use the encoder network to represent the image and a left-to-right decoder to generate the caption text. The standard auto-regressive language model objective is used to maximize the data probability. These two tasks can help learn high-quality visual representations~\citep{zhang2021vinvl, desai2020virtex}. Detection task can learn object-level visual semantics, while the image caption task can capture text-aligned visual semantics. These two kinds of visual semantics matter significantly in VLP cross-modal fusion. During fine-tuning, E2E-VLP can be flexibly applied to vision-language understanding tasks with the encoder module, and vision-language generation tasks with the encoder-decoder module.



We evaluate E2E-VLP on a variety of representative vision-language tasks, including visual question answering, natural language visual reasoning, cross-modal retrieval and image captioning. With the new end-to-end pre-training paradigm, we can obtain surprising good performance across different V+L tasks and greatly decrease the online inference time with the new one-stage solution.

We make the following major contributions in this paper: 
\begin{myitemize2}
\itemsep0em
\item We propose the first end-to-end vision-language pre-trained model for both V+L understanding and generation, namely E2E-VLP, which can achieve comparable or superior performance with faster online inference speedup.
\item E2E-VLP is the first model that incorporates fine-grained visual pre-training in an encoder-decoder architecture, which paves a new way for designing advanced vision and language pre-training tasks.
\item We enhance cross-modal feature fusion by visual learning of object detection and image caption, which has empirically shown to be effective for vision-language pre-training.
\end{myitemize2}

\begin{figure*}
\centering
\includegraphics[width=0.9\textwidth]{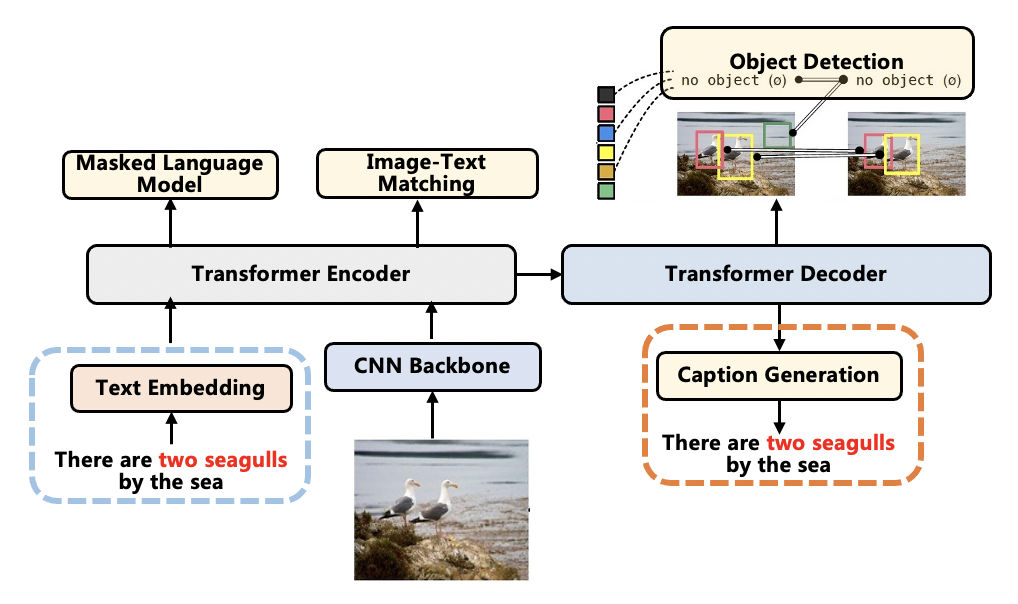}
\caption{The overall framework of E2E-VLP. Our model employs a unified encoder-decoder transformer framework to learn visual representation, and semantic alignment between image and text jointly. 
} 

\label{fig:framework} \vspace{-2mm}
\end{figure*}
\section{Related Work}
Self-supervised pre-training has substantially advanced the performance across a variety of natural language understanding ~\citep{devlin2018bert, liu2019roberta, wang2019structbert, lan2019albert} and text generation tasks~\citep{song2019mass, lewis2019bart, bi2020palm}. Inspired by language model pre-training, several researchers propose Vision-language pre-training(VLP) models on large-scale image-text pairs, which  has proved effective for a wide range of vision-language (VL) tasks, such as VQA~\citep{antol2015vqa}, NLVR~\citep{young2014image}, Cross-modal Retrieval~\cite{suhr2018corpus}. 

The current VLP models mainly take two-step training pipeline, which consists of extracting semantic visual features by object detector and training the cross-modal pre-training model to align text and visual features. In this kind of method, there are mainly two broad directions to conduct vision-language pre-training. The first line uses a single-stream transformer architecture ~\cite{vaswani2017attention} to model both image and text representations in a unified semantic space such as VLBERT~\citep{su2019vl}, UNITER ~\citep{chenuniter} and OSCAR~\citep{li2020oscar}. In contrast, the other line adopts a two-stream Transformer architecture that first encodes the image and text modalities separately, and then fuses the cross-modal representations with another Transformer network, such as LXMERT~\citep{tan2019lxmert} and ERNIE-ViL~\citep{yu2020ernie}. Besides, SemVLP~\citep{li2021semvlp} is pre-trained iteratively with two prevalent fashions. These methods are directly trained and evaluated on cached visual features, which imposes unnecessary constraints on model designs and makes it hard to enable an end-to-end vision-language pre-training. Furthermore, Pixel-BERT ~\citep{huang2020pixel} represents the first and only work to pre-train with grid features in an end-to-end fashion. However, due to the characteristics of learnt grid features, the end-to-end pre-training is conducted without object-level visual tasks, which is important in aligning the semantics between cross-modal representations.

In this paper, we focus on enhancing the end-to-end vision-language pre-training with more fine-grained visual semantic learning. The object detection task and image caption task are incorporated into the pre-training stage for further improving the fine-grained visual-language understanding and generation abilities.

\section{E2E-VLP Pre-training}
\subsection{Model Architecture}
The architecture of E2E-VLP is shown in Figure~\ref{fig:framework}. Inspired by the recent breakthrough of using Transformer on computer vision tasks such as DETR~\citep{carion2020end} and ViT Transformer~\citep{dosovitskiy2020image}, we propose to use a Transformer encoder-decoder framework~\citep{vaswani2017attention} for cross-modal learning, and a simple CNN backbone module is used as the image encoder for extracting visual representations from pixels so as to allow for more flexible network design. We jointly train the whole framework in an end-to-end fashion, so as to learn the generic visual representations and high-level cross-modal alignment simultaneously. Different V+L pre-training tasks are designed to further enhance the cross-modal understanding and generation abilities.
Next, we describe each component of this model in detail.

\subsubsection{Input Representations} \label{sec:3}
The input to E2E-VLP is an image and its related text (e.g. caption text). We first introduce the way to represent the text sequence and raw image pixels as input to the Transformer. 

\paragraph{Sentence Embeddings}
Each sentence is first split into a sequence of sub-words $\{w_{1},...,w_{m}\}$ by WordPiece tokenizer. Then, similar to BERT~\citep{devlin2018bert}, each token $w_i$ is assigned three kinds of embeddings: token, segment and position embeddings. The three embeddings are summed and layer-normalized to represent input sentence representations as a sequence of embedding vectors $E_{emb}=\{e_{CLS},e_{1},...,e_{m},e_{SEP}\}$, where $[CLS]$ and $[SEP]$ are special tokens in BERT.

\paragraph{Image Representations}
For image feature representation, the most existing VLP models follow Bottom-Up and Top-Down Attention~\citep{anderson2018bottom} to extract region features by Faster R-CNN~\citep{ren2015faster} trained on Visual Genome dataset. The detector extracts region features by first detecting regions under pre-defined categories, and then uses the features before the final classifier as the output. These methods are limited to the task-specific visual representation of the specific object detector, which may hinder the generic cross-modal understanding. 

To improve the generalization of the image representation, we learn from pixels to represent an image instead of using bounding boxes. The pixel features are learned by a CNN visual backbone such as ResNet~\cite{he2016deep}. Starting from the initial image $v_{img}\in R^{3\times H_{0}\times W_{0}}$ (with 3 color channels), a conventional CNN backbone generates a lower-resolution activation map $f_{img}\in R^{C\times H\times W}$ using the typical values as in DETR~\citep{carion2020end}: $C=2048$ and $H=\frac{H_0}{32}, W=\frac{w_0}{32}$. Then, we take a $1\times1$ convolution to reduce the channel dimension of the high-level activation map $f$ from $C$ to a smaller dimension $d$, creating a new feature map $z_{img}\in R^{d\times H\times W}$. The encoder expects a sequence as input, hence we collapse the spatial dimensions of $z_{img}$ into one dimension, resulting in a $HW\times d$ feature map $Z_{img}$.  Since the transformer architecture is permutation-invariant, we supplement the feature maps with fixed positional encodings~\citep{parmar2018image} that are added to the input of each attention layer. Finally, the sequential image representation $Z_{img}=\{o_1,...,o_{HW}\}$ can be seen as a $HW$ length of $d$-dimensional vector. 

\subsubsection{Cross-modal Encoder Pre-training}

Given the embeddings of the tokens for the sentence $\{e_i\}_{i=1}^m$ and the sequential image representations $\{o_j\}_{j=1}^n$, we adopt the Transformer encoder to learn cross-modal attention between image grid features and language tokens. The encoder is a stacked model with $L$ standard blocks, where the $l$-th block consists of a
multi-head self-attention module and a feed forward network (FFN). To allow a fine-grained feature-level semantic fusion, we directly concatenate the derived image features and text embeddings to construct the input sequence, which is formulated as: 
$\{e_{CLS},e_{1},...,e_{m},e_{SEP},o_1,...,o_{HW}\}$.

The CNN backbone for visual representation learning and the Transformer for cross-modal semantic fusion is combined into a single model, which is end-to-end trainable. In this way, the learnt visual feature representation can be more suitable for the pre-training tasks of generic cross-modal understanding. To facilitate cross-modal understanding, we follow~\citep{tan2019lxmert,chenuniter,huang2020pixel} and conduct two popular pre-training tasks in encoder side, including Masked Language Modeling (MLM) and Image-Text Matching (ITM). 
\paragraph{Masked Language Modeling} The task setup is basically the same as in BERT~\citep{devlin2018bert}, we randomly mask 15\% tokens in the text and the model is asked to predict these masked words with the output text and visual representations. Different from MLM task in BERT that only relies on the surrounding text of textual modality for prediction, the masked words will be predicted with the help of image feature map from visual modality so as to resolve ambiguity.

\paragraph{Image-Text Matching}
We randomly sample 50\% mismatched image-text pairs and 50\% matched pairs, and train an classifier to predict whether an image and a sentence match each other on the representation of token [CLS] in the last encoder layer $h^L_{CLS}$.

\subsubsection{Visual-enhanced Decoder}
Due to that the CNN feature map has no object-level semantics, it is difficult to directly align the cross-modal semantics between CNN feature map and the language embeddings. Therefore, we further add a Transformer decoder to help capture the fine-grained semantics of the visual features, where two specific pre-training tasks of object detection and image-caption generation are incorporated.

The decoder adopts the standard architecture of the transformer with multi-headed self-attention followed by cross-attention and a feed forward network (FFN). Both tasks share the same attention parameters of decoder, while using different linear head for the two tasks. The object detection task focuses more on understanding the fine-grained object information within image, while image captioning task helps guide the learning of visual features regarding the textual semantics. 

\paragraph{Enhanced by Object Detection}  Following the one-stage detection model DETR \citep{carion2020end}, we define object detection task as the direct set prediction problem, and use a set-based global loss that forces unique predictions via bipartite matching with the Transformer encoder-decoder architecture. 

Let us denote by $y$ the ground truth set of objects and $\hat{y}=\{\hat{y}_i\}^N_{i=1}$. The set-based loss of bipartite matching is to search for a permutation of $N$ elements $\sigma \in \mathscr{L}_N$ with the lowest cost:
\begin{align}
\hat{\sigma} = \mathop{\arg\min}_{\sigma\in \varphi_{N}}\sum_{i}^{N} \mathcal{L}_{match}(y_i, \hat{y}_{\sigma(i)})
\end{align}
where $\mathcal{L}_{match}(y_i, \hat{y}_{\sigma(i)})$  is a pair-wise matching cost between ground truth $y_i$ and a prediction with index $\sigma(i)$.

The Hungarian algorithm~\citep{stewart2016end} is used to efficiently compute the optimal assignment.  Different from the original DETR for single-modal learning, our cross-modal pre-training with object detection differs in two aspects.

In encoder side, we combine both the visual representation and language embedding as input and reuse the Transformer encoder for cross-modal fusion. In decoder side, we take the learned positional embeddings as the input to multiple $L$ Transformer decoder layers, and detects the $N$ objects in parallel at each decoder layer. In addition to the tasks of box coordinate regression and class category prediction, we also incorporate an object attribute prediction task for Visual Genome Dataset so as to enhance the learning of fine-grained semantics. The model is trained with a negative log-likelihood loss for attribute, class prediction and a box regression loss defined as follows:
\begin{align*}
 \mathcal{L}_{v}(y,\hat{y}) =& \sum_{i=1}^{N}[-{\rm log} \hat{p}_{\hat{\sigma}(i)}(a_i) -{\rm log} \hat{p}_{\hat{\sigma}(i)}(c_i) +\mathscr{1} \\ &+ \mathcal{L}_{box} (b_{i}, \hat{b}_{\hat{\sigma}(i)}(i))] 
\end{align*}
where $\hat{p}_{\hat{\sigma}(i)}(a_i), \hat{p}_{\hat{\sigma}(i)}(c_i)$ is the attribute and class probability, $\mathcal{L}_{box} (b_{i}, \hat{b}_{\hat{\sigma}(i)}(i))$ is a normalized bounding boxes regression loss as in~\citep{carion2020end}. 

\paragraph{Enhanced by Image Captioning}
To guide the learning of visual features in regards to the textual semantics, we use semantically dense captions to learn vision representations with sequence-to-sequence (Seq2Seq) image-to-text generation task.
The decoder is pre-trained to auto-regressively generate the target text based on the contextual representations from the image encoder. The pre-training loss for the decoder is defined as:
\begin{equation} \label{equ:2}
        \mathcal{L}_{dec}=-\sum_{(x,y)\in (\mathcal{X},\mathcal{Y})}\log \prod_{t=1}^n P(y_t|y_{<t},x)
\end{equation}
where $\mathcal{X}$ represents the sequence of vision context, $\mathcal{Y}$ represents the set of text to be generated and $n$ is the length of tokens in output text $y$.

\subsection{Joint Training}
We pre-train E2E-VLP with all the encoder and decoder pre-training tasks (i.e., Masked Language Modeling, Image-Text Matching, Object Detection, Image-to-Text Generation) jointly by minimizing the four loss functions as:
\begin{equation} \label{equ:3}
        \mathcal{L}=\mathcal{L}_{mlm}+\mathcal{L}_{itm}+\mathcal{L}_{v} + \mathcal{L}_{dec}
\end{equation}

\section{Experiments}
\subsection{Pre-training Dataset}
We pre-train our E2E-VLP on two in-domain image-text datasets: MS-COCO~\citep{lin2014microsoft} and Visual Genome \citep{krishna2017visual}. We utilize the object detection and image caption annotations in MS-COCO, and object detection, region description annotations in Visual Genome. The total amount of the dataset is 6.01M image-and-sentence pairs on 180K distinct images.
\subsection{Implementation Details}
The maximum sequence length for the sentence is set as 40. We use scale augmentation, and resize the input images so that the shortest side is at least 480 and at most 800 pixels while the longest is at most 1333 \cite{carion2020end}.
For the model architecture, we pre-train E2E-VLP with 6 and 12 layers of Transformer encoder respectively, while the decoder is fixed as 6 layers. Each layer block has 256 hidden units and 12 self-attention heads, the intermediate layer size is 1,024. The visual backbone is selected as ResNet with different sizes \citep{he2016deep} from torchvision with frozen batch-norm layers. We pre-train E2E-VLP model with a total batch size of 32 for 200 epoches on 8 V100 GPUs. We use the AdamW optimizor~\citep{loshchilov2018decoupled} for both the Transformer and ResNet. The initial learning rate is set as   $10^{-4}$ for Transformer and $10^{-5}$ for ResNet. The weight decay is set as $10^{-4}$.

\section{Experiments}
\begin{table*}
\centering
\small
\begin{tabular}{cl|c|ll|ll|ll}
\toprule
\multicolumn{2}{c|}{\multirow{2}{*}{Models}}      &
\multirow{2}{*}{Params} &
\multicolumn{2}{c|}{VQA} & \multicolumn{2}{c|}{NLVR2} & \multicolumn{2}{c}{COCO Caption}  \\
\multicolumn{2}{l|}{}    &    & Test-dev & Test-std    &  Dev       &Test-P  &   BLEU4   & CIDEr             \\
\midrule
\multirow{5}{*}{Single-stream} & VisualBERT &  110M    & 70.80    & 71.00        & -     & -                & -     & -                 \\
            & VLP & 110M       & 70.5    & 70.7        & -     & -    & 36.5                & 116.9 \\
                  & VLBERT & 110M   & 71.16    & -        & - & -            & - & -              \\
			      & Unicoder-VL & 110M & - & - & - & -  & - & -  \\
                              & UNITER & 110M     & 72.70    & 72.91        & 77.14 & 77.87             & - & -             \\
                              & OSCAR & 110M       & 73.16    & 73.61        & 78.07     & 78.36     & 36.5                & 123.7                         \\
\midrule
\multirow{4}{*}{Two-stream}    & ViLBERT & 221M & 70.55    & 70.92        & 67.40 & 67.00             & -          & -                 \\
                              & 12-in-1 & 221M & 73.15    & -        & - & - & -            & -                     \\
                              & LXMERT & 183M          & 72.42    & 72.54        & 74.90     & 74.50     & -                        & -                 \\
                              & ERNIE-ViL & ~210M  & 72.62    & 72.85        & - & - & -          & -           \\
\midrule
End2End & PixelBERT & 142M &71.35 &71.42 & 71.7& 72.4&- &- \\
\midrule
Our  Model             & E2E-VLP & 94M &  73.25   &   73.67    &77.25       &     77.96  &         36.2      &       117.3          \\
\bottomrule
\end{tabular}
\caption{Evaluation Results on VQA, NLVR2 and Image Caption.}
\label{table:overall1}
\end{table*}
 
\begin{table*}[!htb]
\centering
\small
\begin{tabular}{cl|c|lll|lll}
\toprule
\multicolumn{2}{c|}{\multirow{2}{*}{Models}}      &
\multirow{2}{*}{Params} &
 \multicolumn{3}{c|}{IR-Flickr30K} & \multicolumn{3}{c}{TR-Flickr30K}  \\
\multicolumn{2}{l|}{}   &              &             R@1   & R@5   & R@10             & R@1   & R@5   & R@10              \\
\midrule
\multirow{5}{*}{Single-stream} & VisualBERT &  110M        & -     & -     & -                & -     & -     & -                 \\
                  & VLBERT & 110M   & - & - & -            & - & - & -             \\
			      & Unicoder-VL & 110M & 71.50 & 90.90 & 94.90 & 86.20 & 96.30 & 99.00 \\
                               & UNITER & 110M      & 72.52 & 92.36 & 96.08            & 85.90 & 97.10 & 98.80             \\
                               & OSCAR & 110M        & -     & -     & -                & -     & -     & -                 \\
\midrule
\multirow{4}{*}{Two-stream}    & ViLBERT & 221M        & 58.20 & 84.90 & 91.52            & -     & -     & -                 \\
                               & 12-in-1 & 221M & 67.90 & - & -            & -     & -     & -                 \\
                               & LXMERT & 183M               & -     & -     & -                & -     & -     & -                 \\
                               & ERNIE-ViL & ~210M   & 74.44 & 92.72 & 95.94            & 86.70 & 97.80 & 99.00             \\
\midrule
End2End & PixelBERT & 142M & 59.8 & 85.5 & 91.6            & 75.7 & 94.7 & 97.1         \\
\midrule
Our  Model             & E2E-VLP & 94M      &  73.58    & 92.42    &    96.03             & 86.24     & 97.50     & 98.92               \\
\bottomrule
\end{tabular}
\caption{Evaluation Results on Flickr30K.}
\label{table:retrieval}
\end{table*}


\subsection{Downstream Tasks}
We compare E2E-VLP model against other competitive VLP models of the comparable model size on the following downstream V+L tasks.
\begin{myitemize2}
\itemsep0em
\item \textbf{VQA v2.0}~\cite{antol2015vqa}: The VQA task requires the model to answer natural language questions given an image. We conduct experiments on the widely-used VQA v2.0 dataset~\citep{antol2015vqa}, which contains 204K images and 1.1M questions about these images. Following~\citep{anderson2018bottom}, we treat VQA as a multi-label classification task by picking an answer from a shared set consisting of 3,129 answers.
To fine-tune VQA task, we use a binary cross-entropy loss to train a multi-label classifier, we train
with a batch size of 32 for 12 epochs. We set an initial learning rate of 1e-4 which decays by 0.1 at the end of epoch 6 and epoch 9.

\item \textbf{NLVR2}~\cite{suhr2018corpus}: NLVR2~\citep{suhr2018corpus} is a challenging task for visual reasoning. The goal is to determine whether a natural language statement is true about a pair of images. It consists of 86K/7K data for training/development. Since each data example in NLVR2 has two natural images $img_0$, $img_1$ and one language statement $s$, we concatenate the given sentence and each image to build two sequences, and then train a binary classifier based on the concatenation of the two outputs. We fine-tune NLVR model with a batch size of 32 for 12 epochs, and set an initial learning rate of 1e-4 which decays by 0.1 at the end of epoch 6 and epoch 9.


\item \textbf{Image Caption}: A visual generation task that requires the model to generate the content of an image. To fine-tune Image Caption task, we use the seq2seq loss with label smoothing\citep{szegedy2016rethinking}. During inference, we use beam search (i.e., beam size=4), and set  $\alpha=0.9$ for the length penalty~\citep{wu2016google}. We set initial learning rate of 1e-4 which decays by 0.1 at the end of epoch 6 and epoch 9. We report our results on the COCO image captioning dataset~\citep{chen2015microsoft}.

\item \textbf{Image-Text Retrieval}: The image-text retrieval task consists of two sub-tasks: image retrieval and text retrieval, depending on which modality is used as the retrieval target. We conduct experiments on Flickr30K dataset~\citep{young2014image}, which contains 31,000 images collected from Flickr website and each image has 5 captions. We follow the same split in~\citep{lee2018stacked} for training and evaluation. During fine-tuning, we follow the method in UNITER~\citep{chenuniter} and formulate it as a ranking problem. We use the hidden state of $h^L_{CLS}$ to compute the similarity scores for the sampled positive and negative pairs, and maximize the margin between them through circle loss~\citep{sun2020circle} as ERNIE-ViL~\citep{yu2020ernie}. We fine-tune our model with a batch size of 64 and a learning rate of 5e-5 for 4 epochs.

\end{myitemize2}
\subsection{Baseline Methods}
We compare our E2E-VLP model with all the three prevalent VLP architectures: i.e., single-stream and two-stream architectures of two-step pipeline framework and end-to-end one-step solution. Single-stream architecture uses a unified Transformer to encode the vision-language inputs, including the state-of-the-art methods such as OSCAR\citep{li2020oscar}, UNITER\citep{chenuniter}, Unicoder-VL~\citep{li2020unicoder}, VLBERT~\citep{su2019vl} and VLP~\citep{zhou2020unified}. Image and text are separately encoded firstly and then fused together in two-stream architecture, including the state-of-the-art methods such as ERNIE-VIL\citep{yu2020ernie}, LXMERT~\citep{tan2019lxmert}, ViLBERT~\citep{lu2019vilbert,lu202012}. These two architectures both adopt the region-based visual features, where a object detector is first used to obtain the object-level feature representations. We also compare with the only end-to-end solution PixelBERT~\citep{huang2020pixel}. PixelBERT adopts a random pixel sampling strategy to conduct the cross-modal pre-training, while it has no visual semantic understanding tasks for pre-training which is very important in V+L tasks.

\subsection{Main Results}
The results on the downstream V+L tasks are shown in Table \ref{table:overall1}. It can be observed that: 1) with less parameters and only in-domain pre-training data (MS-COCO and Visual Genome), E2E-VLP can consistently achieve comparable performance against two-step region feature-based methods such as OSCAR and ERNIE-VIL. It shows the effectiveness of our end-to-end grid feature-based method, which can offer new perspectives to address the cross-modal pre-training and conduct fusion at a more fine-grained level. It has the potential of removing the complex procedure of region feature extraction, and facilitate deeper interaction between visual feature and text data in an end-to-end fashion. 2) Our E2E-VLP method can significantly improve upon the end-to-end method  PixelBERT, which demonstrates the advantages of our method for enhancing the fine-grained visual learning with object detection and image captioning,
 



\begin{table}
\centering
\small
\begin{tabular}{l|c|c} 
\toprule
Model &VQA  & NLVR2 \\
\midrule
E2E-VLP & 70.76 &  72.12\\
-Image-to-Text Generation & 70.20 & 71.59  \\
-Attribute Prediction & 69.92 & 70.92\\
-Object Detection & 68.85 & 70.38  \\
\bottomrule
\end{tabular}
\caption{Ablation tests for different visual pre-training tasks of E2E-VLP (6 layer encoder, and ResNet50 backbone) on development set.}
\label{table:ablation} 
\end{table}

 
\subsection{Importance of Visual Learning}
To further investigate the importance of each component in our method, we conduct ablation studies to assess the impact of different visual learning tasks on the VQA and NLVR2 development set. Table \ref{table:ablation} shows the result. We can see that: 1) all the three visual pre-training tasks contribute to the final performance gain, and removing each of them can decrease the performance on both tasks. The object detection and attribute prediction tasks can help capture fine-grained object-level semantics within the image, which is consistent with the previous two-step solutions that using region features from the detection can help improve the performance for cross-modal understanding. The image-to-text generation task can help guide the learning of visual features in regards to the textual semantics, which has the same conclusion as VirTex~\citep{desai2020virtex}. 2) Among the different visual pre-training tasks, the Object Detection and Attribute Prediction tasks are more important than the Image-to-Text Generation task, this may be due to the fact that the typical cross-modal downstream tasks such as VQA and NLVR2 focus more on the fine-grained semantics of the objects within image. 

\subsection{Inference Efficiency}
One of the biggest advantages of end-to-end VLP method is the inference efficiency with one single stage. Therefore, we further examine the online inference efficiency of E2E-VLP, compared with the two-step region-based models (UNITER and LXMERT) and the existing end-to-end VLP model (PixelBERT). We examine the average inference time (per query) of different models on the VQA dataset. The result is shown in Table~\ref{table:speed}. We can see that: 1) the end-to-end methods can be much more efficient in online inference (2-3 times speedup) than the two-step model. We further analyze the inference time of different components of two-step models and find that among the total cost of 500ms per image-text pair, about 80\% of the total time is used to extract region-based features using Faster R-CNN~\citep{ren2015faster}. It takes much time for region selection and this will happen twice when extracting the final regions, and it contains many complicated post-processing procedures. 2) Our E2E-VLP model can achieve comparable results on both the VQA and NLVR2 datasets by saving about 3.5 times running time. Besides, we can also use a smaller image size to further improving the inference speed. Compared with PixelBERT, E2E-VLP can also obtain some speed-ups due to the reason that the Transformer hidden size of E2E-VLP is only 256, which makes E2E-VLP more light-weight and flexible. Our end-to-end solution can significantly improve the performance upon PixelBERT, because there are no visual pre-training tasks for PixelBERT and we enhance the pre-training of E2E-VLP with both the fine-grained Object Detection and Image Captioning tasks. 



\begin{table}
\centering
\small
\begin{tabular}{c|c|c|c|c} 
\toprule
\multirow{2}{*}{Model} &\multirow{2}{*}{Parameters}& Avg Time &\multirow{2}{*}{VQA}  & \multirow{2}{*}{NLVR2} \\
&&(ms)&&\\
\midrule
LXMERT&183M& 496&72.42&72.54\\
UNITER &110M& 501 &72.70&77.14\\
Pixel-BERT &142M&201&71.35&71.7\\
\midrule
E2E-VLP &94M&192&73.25&77.25\\
\bottomrule
\end{tabular}
\caption{Results of the inference comparison of different pre-trained model architectures on the VQA and NLVR2 dataset.}
\label{table:speed} 
\end{table}

\begin{table}
\centering
\small
\begin{tabular}{c|c|c|c|c} 
\toprule
Layers  & Backbone & Params &VQA  & NLVR2 \\
\midrule
6 & r50& 49M& 70.56& 72.12 \\
6 & r101 &68M & 71.42& 74.34\\
6 & r152 & 84M& 72.23 & 76.21\\
12 & r50 & 59M& 71.34& 73.04\\
12 & r101 & 78M& 72.43& 75.23\\
12 & r152 & 94M& 73.25& 77.25\\
\bottomrule
\end{tabular}
\caption{Results of different pre-trained model architectures on development set.}
\label{table:size} 
\end{table}

\subsection{Architecture Selection}
Since our whole framework contains both the visual backbone and Transformer network as a whole, we further study the importance of different model architectures by changing the number of Transformer encoder layers and the different ResNet visual backbone layers. We expect to further examine whether the visual backbone or Transformer network is more important for the cross-modal understanding and fusion. From Table \ref{table:size}, we can see that both adding more Transformer encoder layers and using more complicated visual backbones can contribute to the final performance gain, which proves the importance of both modules for cross-modal understanding. Learning better visual features and conducting more deeply interacted visual-language fusion are both important for V+L tasks. Besides, we can see that using a more strong visual backbone (such as ResNet 152) can give more benefit to the final performance than just increasing the number of Transformer encoder layers from 6 to 12. This may be due to the fact that visual semantic understanding is rather important in V+L tasks and that is also why we design more fine-grained visual pre-training tasks for further enhancing the learning of E2E-VLP.

\subsection{Impact of Input Image Size}
As mentioned in Section~\ref{sec:3}, the sequence length of the visual features is determined by the image size $HW$. Therefore, the final sequence length of the input to the transformer also largely depends on the image size, which can in turn influence the inference speed of our whole framework. We further analyze the impact of input image size to the efficiency and effectiveness of E2E-VLP. 
The results of E2E-VLP with different image sizes as input are shown in Table \ref{table:img_size}. From the results, we can see that E2E-VLP benefits from larger images as input, and for larger images, the sequence length of the visual representation is longer and more information is embedded in the visual representation. The cross-modal Transformer is capable of learning more fine-grained vision-language fusion for better performance. Moreover, down-sampling the image to a smaller size can significantly improve the inference speed of E2E-VLP model, while the model accuracy only decreases a little. For example, when changing the input size from (800, 1333) to (448, 448), the inference can be about 5 times faster while the performance only decreases about 2\%-3\%.






\begin{table}
\centering
\small
\begin{tabular}{cc|c|c|c} 
\toprule
 \multicolumn{2}{c|}{Input Size} & \multirow{2}{*}{Speedup} & \multirow{2}{*}{VQA} & \multirow{2}{*}{NLVR2}\\
shorter side & longer side & & & \\
\midrule
448& 448& 5x & 71.14 & 75.43\\
448 & 746 & 3x & 72.04 & 75.79 \\
600 & 1000 & 1.5x & 73.08 &76.87\\
800 & 1333 & - & 73.25 &77.25\\
\bottomrule
\end{tabular}
\caption{Impact of input image size on the VQA and NLVR2 set.}
\label{table:img_size} 
\end{table}

\subsection{Object Detection with Paired Text}
\begin{table}
\centering
\small
\begin{tabular}{c|c|c|c|c|c} 
\toprule
Model&AP & $AP_{50}$ &$AP_{S}$ &$AP_{M}$&$AP_{L}$    \\
\midrule
DETR & 40.6 &  61.6 & 19.9 &44.3&60.2 \\
E2E-VLP &41.9 &62.6 & 20.3&45.6&61.1\\
\bottomrule
\end{tabular}
\caption{Results of object detection on MSCOCO development dataset}
\label{table:detection} 
\end{table}

Finally, we expect to further examine whether the cross-modal fusion is stable and E2E-VLP capture fine-grained semantics by visual learning. Therefore, we encode both the image content and caption text with E2E-VLP, and directly fine-tune it on MSCOCO object detection benchmark dataset with the decoder as in DETR\citep{carion2020end}. 
Table~\ref{table:detection} shows the detection result.
We can see that our E2E-VLP model can also support the Object Detection task based on text-image pairs and perform surprising well compared with the original DETR model. This phenomenon may also demonstrate that E2E-VLP well captures the fine-grained semantics within image and can appropriately fuse the multi-modal information for conducting visual-only task. 




\section{Conclusion}
In this paper, we propose a new end-to-end paradigm for pixel-level vision-language pre-training, to jointly learn visual representation, and semantic alignments between image and text. Different from the previous methods using the region features in a two-stage pipeline, we propose to use the more flexible and efficient image grid features for vision-language pre-training. We further incorporate  the  tasks  of  object  detection  and  image captioning into pre-training with a unified Transformer encoder-decoder architecture for enhancing visual learning.  The experiments on well-established vision-language downstream tasks demonstrate the effectiveness and efficiency of our E2E-VLP model. We hope that this study can potentially offer new perspectives and guide for end-to-end vision-language pre-training.

In the future, we will explore more deeply interacted ways for image-text fusion from a bottom layer, and incorporate more advanced vision and language pre-training tasks for further improving the performance.

\bibliographystyle{acl_natbib}
\bibliography{acl2021}


\end{document}